\documentclass[runningheads]{llncs}

\usepackage{lipsum}
\usepackage[export]{adjustbox}
\usepackage{eccv}

\usepackage{makecell}
\usepackage{longtable}
\usepackage{eccvabbrv}
\usepackage[table]{xcolor}
\usepackage{enumerate}
\usepackage{pifont}
\usepackage{wrapfig}  
\newcommand{\cmark}{\ding{51}}  % ✓ checkmark
\newcommand{\xmark}{\ding{55}}  % ✗ cross mark
\usepackage{graphicx}
\usepackage{booktabs}
\usepackage{multirow}
\usepackage[accsupp]{axessibility}  % Improves PDF readability for those with disabilities.
\usepackage{float}

\usepackage[bookmarks=false, linkcolor=blue, urlcolor=blue, citecolor=blue]{hyperref} 
\hypersetup{
    colorlinks=true,
    linkcolor=red,
    filecolor=magenta,      
    urlcolor=blue,
    pdfstartview={FitH},
    citecolor =blue
    }

\usepackage{orcidlink}

\begin{document}

% ---------------------------------------------------------------
% TODO REVIEW: Replace with your title
\title{Gripper-aware Vision Language Action Models} 

% TODO REVIEW: If the paper title is too long for the running head, you can set
% an abbreviated paper title here. If not, comment out.
\titlerunning{Gripper-aware Vision Language Action Models}

% TODO FINAL: Replace with your author list. 
% Include the authors' OCRID for the camera-ready version, if at all possible.
\author{Hanyi Zhang\inst{1}\orcidlink{0009-0009-5141-2789} \and
Zihong Luo\inst{1}\orcidlink{0009-0008-0421-5873} \and
Tianyu Li\inst{1} \and 
Khang Nguyen \inst{2} \and 
Basu Hela \inst{3}\orcidlink{0009-0000-0211-9679} \and
Shreyas Kumar \inst{3} \and
Ngoc Duy Tran \inst{3} \and
Feng Dai \inst{4} \and
Charith Munasinghe \inst{5} \and
Jorge Peña Queralta \inst{5} \and
Giovanni Toffetti \inst{5}\orcidlink{00000-0002-0740-1355} \and
Khoa Vo \inst{6}\orcidlink{0000-0003-0277-7094} \and
Ngan Le \inst{6} \and
Ravi Prakash \inst{3} \and
Quan Vuong \inst{7} \and 
Tung D. Ta \inst{4} \and
Long Hu \inst{8} \and
Anh Nguyen\inst{1} \orcidlink{0000-0002-1449-211X} \and
Baoru Huang\inst{1}\orcidlink{0000-0002-4421-652X}
}
% \institute{University of Liverpool, Liverpool L69 3BX, UK \and
% Mohamed bin Zayed University of Artificial Intelligence, Abu Dhabi, UAE \and
% Indian Institute of Science, Bangalore 560012, India \and
% The University of Tokyo, Tokyo 113-8656, Japan \and
% Zürcher Hochschule für Angewandte Wissenschaften (ZHAW), Winterthur, Switzerland \and
% University of Arkansas, Fayetteville, AR 72701, USA \and
% Physical Intelligence, San Francisco, CA, USA \and
% Huazhong University of Science and Technology, Wuhan 430074, China
% TODO FINAL: Replace with an abbreviated list of authors.
\authorrunning{H.~Zhang et al.}
% First names are abbreviated in the running head.
% If there are more than two authors, 'et al.' is used.

% TODO FINAL: Replace with your institution list.
\institute{University of Liverpool, UK \and
Mohamed bin Zayed University of Artificial Intelligence, UAE \and
Indian Institute of Science, India \and
The University of Tokyo, Japan \and
Zürcher Hochschule für Angewandte Wissenschaften, Switzerland \and
University of Arkansas, USA \and
Physical Intelligence, USA \and
Huazhong University of Science and Technology, China
}

\maketitle
\begin{center}
\vspace{-5ex} 
  \small \url{https://airvlab.github.io/G-VLA/}
\end{center}
\vspace{-6ex} 
\begin{figure*}[!h]
    \centering
    \includegraphics[width=\textwidth]{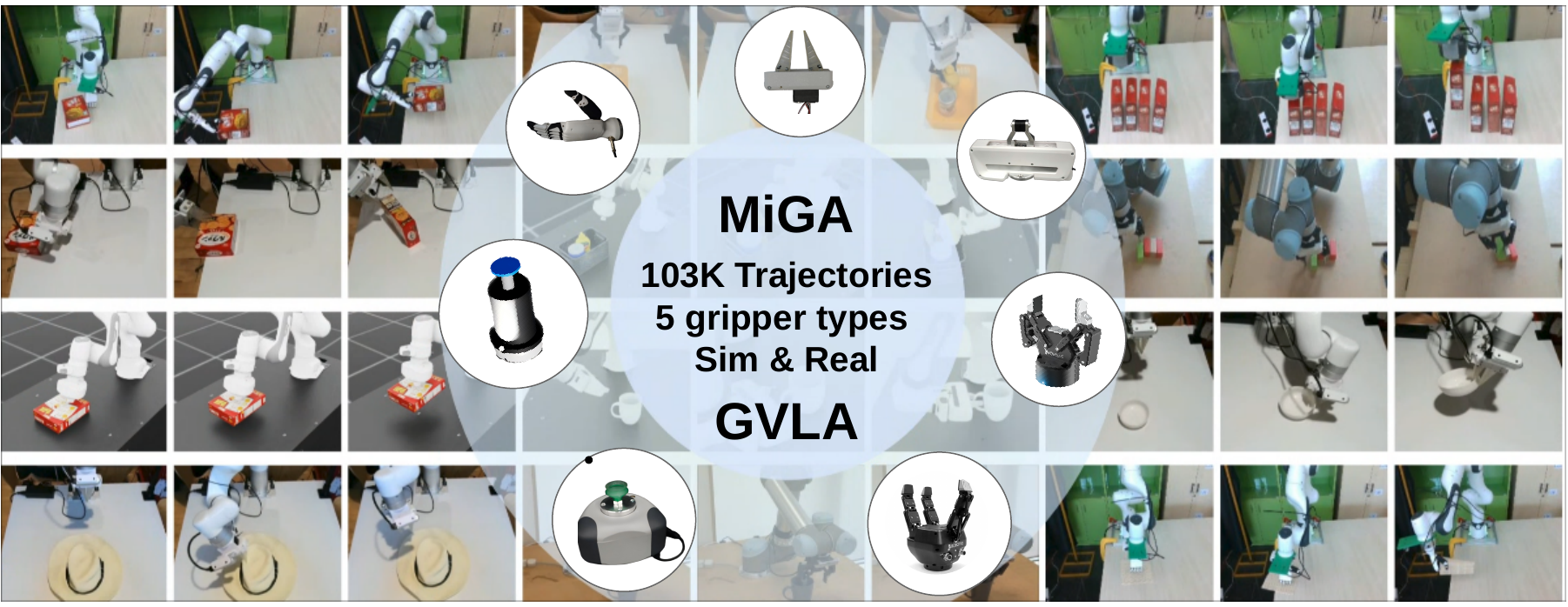} % Updated to use the provided diagram
    \vspace{-0.5cm}
    \caption{We tackle the challenge of gripper-specific grasping in robotics, e.g., a suction cup gripper lifts a thin, flat, DVD-case-like box from above, while a parallel-jaw gripper pushes it to the table edge and grasps it from the side. We first introduce MiGA, a large-scale multi-gripper-aware dataset collected by diverse gripper types. Leveraging this dataset, we then develop a Gripper-aware Vision Language Action (GVLA) model that incorporates gripper conditioning through a new multi-gripper tokenization method and a dual Mixture-of-Adapter. Our approach improves task performance over gripper-agnostic baselines and enables cross-gripper adaptation.}
    \label{fig:intro}
\end{figure*}
\vspace{-5ex}  
%%%%%%%%% ABSTRACT
\begin{abstract}
\vspace{-3ex}
Vision language action models (VLAs) have advanced general purpose robotic grasping and manipulation by enabling robots to interpret visual observations and natural language instructions to generate executable action sequences. However, existing VLAs often implicitly assume gripper invariance, despite grasping strategies being inherently embodiment-dependent. Different gripper types, such as parallel-jaw and suction, usually require distinct interaction strategies to achieve the same grasping objective. Moreover, current datasets for VLAs predominantly rely on parallel-jaw grippers, limiting gripper-aware learning.
To address this gap, we introduce MiGA, a multi-gripper-aware dataset spanning five distinct gripper types across multiple robots with 103,000 demonstrations, explicitly capturing strategy divergence under shared task objectives. We further propose GVLA, which combines a new multi-gripper tokenizer with adapter-based policy routing. Our new gripper encoding induces structured embedding information that balances parameter sharing and strategy differentiation, while layer-wise probing confirms meaningful gripper-conditioned representations for VLAs. Intensive experiments in both simulation and real-world robots show that our GVLA outperforms the current baselines across evaluated settings. Our method also improves zero-shot generalization or few-shot adaptation to new objects or unseen tasks, and enable more efficient gripper adaptation.

% with familiar grippers and enables rapid few-shot adaptation to novel grippers and objects.

%\href{https://drive.google.com/file/d/18rn1P9SOD3mq35F1RaPBAbxLDy82plcz/view?usp=sharing}{draw.io for all figures}

\end{abstract}

% draw.io
% introduction figure

% dataset distribution:
% https://drive.google.com/file/d/1dABTE9wsm7wAQ6qJ5kIGD9be78xYwK7w/view?usp=sharing

%%%%%%%%% BODY TEXT
% \textbf{1st paragraph: robotics grasping introduction and their current limitation - NO ``COMPLEX TASKS"} 

\section{Introduction}
\label{sec:introduction}
\vspace{-2ex}
General-purpose robotic grasping and manipulation with vision-language-action models (VLAs) have been making a transformative impact on the robotics research community recently~\cite{vuong2024language, nguyen2023open, stone2023open, deng2025graspvla}. Although achieving promising results, current VLAs implicitly assume gripper invariance across the learning tasks, while gripper configurations and grasping strategies are inherently embodiment-dependent~\cite{zhou2025vacuumvla, zhong2025dexgraspvla, li2025scalable}. Different types of grippers necessitate fundamentally distinct manipulation and grasping strategies. For example, as shown in Fig.~\ref{fig:intro} and our demonstration video, grasping a thin, flat object (\textit{i.e.}, a DVD or a facing-down box) with parallel-jaw grippers may require first repositioning the object by sliding it to the table edge and then grasping it from the side; meanwhile, a suction gripper can directly approach from above and lift it. This example exemplifies a critical insight: achieving the same task objective does not imply a shared strategy space, where successful execution depends on the robot's embodiment and its corresponding feasible interactions. Moreover, incorporating robot hardware configurations into the learning of task space under different task contexts requires VLAs to learn distinct strategies for different gripper embodiments. These expectations for VLAs pose challenges in both dataset collection and the design of learning mechanisms. Herein, we pose the central question: ``\textit{Can VLAs learn embodiment-dependence and strategy-level divergence of multiple robotic grippers to tackle diverse, real-world tasks?}''

% \textcolor{red}{stories to both dataset and technique contributions -- drafting: (1) datasets are embodiment-specific (morphology of grippers, robot platforms, etc.), thus collecting an intensive dataset requires task design and hardware resources, and (2) learning mechanism with diverse demonstrations as well as robot platforms; how to use the features for learning? Also, analyze the drawbacks of each.} 

% \textcolor{red}{What are the advantages of our proposed dataset and method? (1) Dataset: gripper diversity, multiple robot platforms, gripper-binded solutions, substep decomposition, domain of data collection, teleop or operated by humans; and (2) Technique: }

To enable VLAs training, an important requirement is the availability of large-scale datasets. Current robotic datasets, such as Open X-Embodiment~\cite{o2024open}, Bridge V2~\cite{walke2023bridgedata}, and DROID~\cite{khazatsky2024droid} dominantly use parallel-jaw grippers, leaving gripper diversity underexplored~\cite{wen2025gr}. This is increasingly limiting as different grippers such as suction cups, multi-finger hands, and soft grippers are also being used in various robotic tasks~\cite{zhou2025vacuumvla, zhong2025dexgraspvla,li2025scalable}. Moreover, different gripper types often require their own manipulation strategies, while existing datasets do not explicitly encode such embodiment-dependent strategy divergence, preventing VLAs from learning gripper-aware policies. To address this gap, we introduce \textbf{MiGA}, a large-scale multi-gripper-aware dataset comprising $103,000$ demonstrations spanning over various tasks with five distinct gripper types. Unlike prior datasets that provide single-solution trajectories with limited gripper types, MiGA explicitly captures how identical task objectives necessitate fundamentally different strategies across different gripper embodiments. Each task includes demonstrations from at least three different gripper types, yielding $132$ distinct gripper-strategy pairs with natural language descriptions. Collected across multiple robot platforms in both simulation and real-world environments, MiGA provides the first comprehensive dataset for training and benchmarking gripper-aware policies that can reason about gripper-dependent grasping.

\begin{figure}[t]
\centering
    \begin{subfigure}[b]{0.22\textwidth}            
            \includegraphics[width=\textwidth]{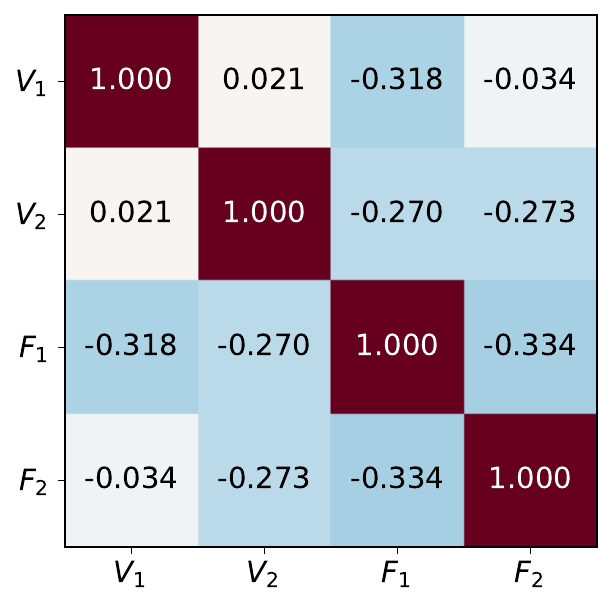}
            \caption{MLP}
            \label{fig:mlp}
    \end{subfigure}%
    \quad
    \begin{subfigure}[b]{0.22\textwidth}
            \centering
            \includegraphics[width=\textwidth]{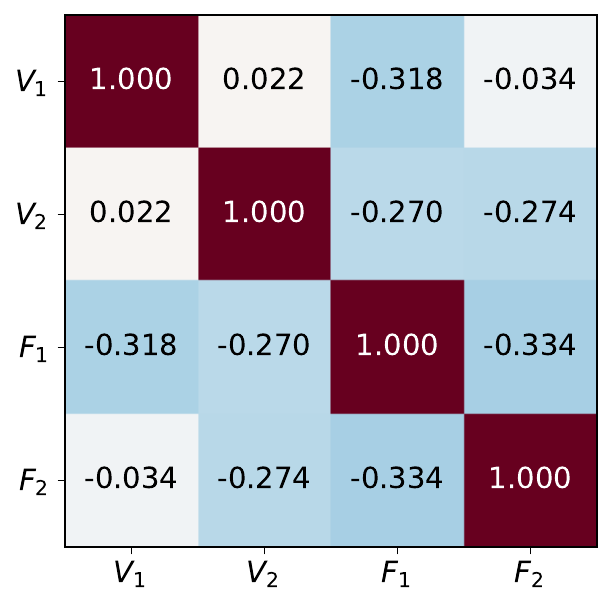}
            \caption{VQ-VAE}
            \label{fig:vqvae}
    \end{subfigure}
    \quad
    \begin{subfigure}[b]{0.22\textwidth}
            \centering
            \includegraphics[width=\textwidth]{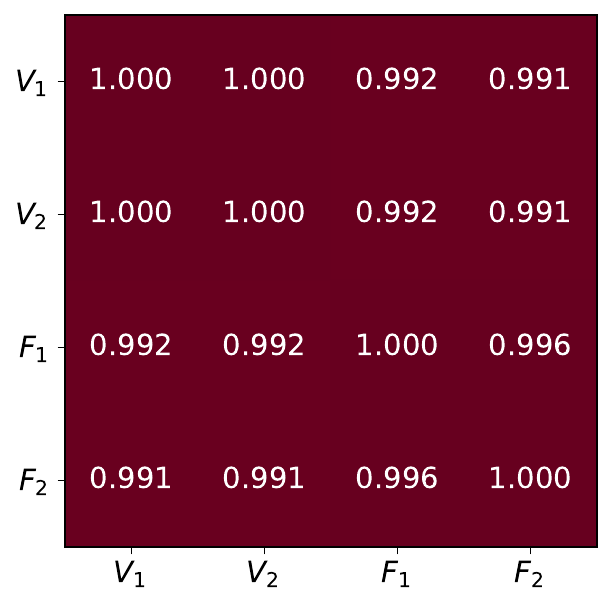}
            \caption{Language Prompt}
            \label{fig:language}
    \end{subfigure}
    \quad
    \begin{subfigure}[b]{0.22\textwidth}
            \centering
            \includegraphics[width=\textwidth]{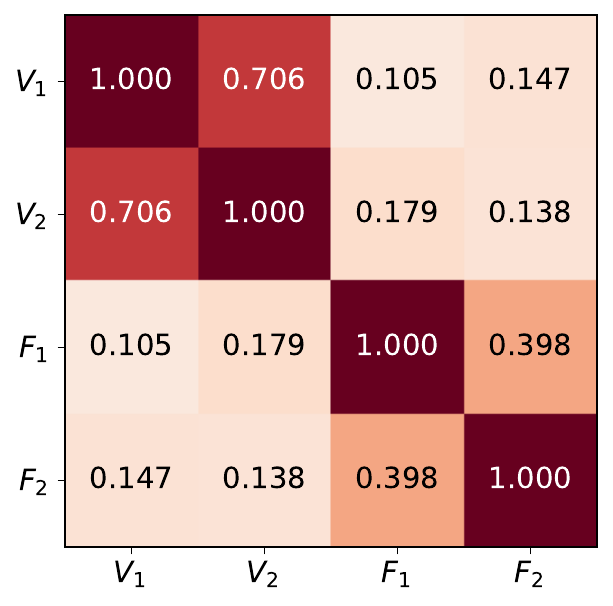}
            \caption{Ours}
            \label{fig:hgp}
    \end{subfigure}
    \vspace{-2ex}
    \caption{Gripper embedding feature heatmap of different gripper representations. Our new multi-gripper tokenization method (d) produces well-clustered representations organized by gripper type. In contrast, other methods (a-c) fail to capture the underlying semantic structure of different gripper types. $V_1$ and $V_2$ denote the vacuum gripper: UR10 Suction Cup~\cite{ur10_suction} and Cobot Pump~\cite{cobot}, $F_1$, and $F_2$ denote finger-based gripper: Franka Panda Hand~\cite{franka2017panda} and Detexterous Hand~\cite{shadowhand}.}
    \label{fig:tokenization}
    \vspace{-5ex}
\end{figure}
Recently, several works have investigated gripper representation for VLAs in robotic applications~\cite{xu2021adagrasp,zhang2025cross,fang2025anydexgrasp}. However, existing gripper representations, such as graph-based encoding~\cite{xu2021adagrasp,khargonkar2024robotfingerprint,yuan2024cross}, primarily target grasp pose transfer among morphologically similar grippers rather than learning robust gripper tokenization. To investigate whether existing gripper representations might suffice, we visualize three gripper tokenization approaches when using the same $\pi_0$~\cite{black2410pi0} backbone: MLP embeddings~\cite{rumelhart1985learning}, VQ-VAE tokenization~\cite{van2017neural}, and language-based embedding~\cite{li2024llava}. Fig.~\ref{fig:tokenization}\textcolor{red}{(a–c)} shows that existing methods struggle to produce meaningful gripper structure: MLP and VQ-VAE produce inconsistent similarities unrelated to morphology, while language prompts collapse to near-identical representations across grippers. %These results indicate that effective gripper representation requires more than distinct embeddings. 
To effectively embed gripper information into VLAs, we introduce two new components: (\textit{i}) a new multi-gripper tokenization that encodes gripper information at three levels: domain-level tokens capturing robot characteristics, type-level tokens encoding gripper category constraints, and instance-level tokens preserving individual gripper features. Fig.~\ref{fig:hgp} shows that our multi-gripper tokenization produces structured representations in the embedding space, with same-type grippers clustering while remaining well separated from different types; and (\textit{ii}) a dual Mixture-of-Adapters (MoA) that directly modulates action generation through gripper-specific expert routing, enabling parameter-efficient fine-tuning while maintaining knowledge sharing across grippers. The intensive experiments on both simulation and real-world robots show that by effectively embedding the gripper information, our method outperforms recent state-of-the-art baselines by $7.62\%$. To summarize, our contributions are as follows:
\vspace{-1ex}
\begin{itemize}
    \item[(1)]We introduce MiGA, a large-scale multi-gripper-aware dataset featuring diverse gripper types on complex tasks, explicitly capturing how the same task requires different strategies depending on the gripper morphology.
    
    \item[(2)]We propose GVLA, a fine-tuning framework with a new multi-gripper tokenizer and MoA to enable strategy-aware manipulation learning.
\end{itemize}

\vspace{-4ex}
\section{Related Work}
\vspace{-2ex}
\textbf{Gripper Representation.} Several robotic research work has recognized that different end-effectors require distinct approaches~\cite{hernandez2023current,d2023multimodal,casas2024multigrippergrasp,song2020grasping,tran2025hybrid}. Existing approaches encode gripper morphology through geometric distance fields between robot and object point clouds~\cite{wei2024d,khargonkar2024robotfingerprint,xu2021adagrasp}, diffusion-based synthesis~\cite{wu2025cedex,nguyen2024lightweight,freiberg2025diffusion,nguyen2024language}, graph-based approaches operate on kinematic topology~\cite{attarian2023geometry,wei2024geomatch++,patel2025get}, and contact-centric representations decouple hand embodiment from object geometry~\cite{li2022gendexgrasp,fang2025anydexgrasp,zhang2025cross}. Beyond these, a subset of works further explores cross-gripper transfer via shared eigengrasp spaces~\cite{yuan2024cross} and latent action alignment~\cite{bauer2025latent}. Despite these advances, existing methods primarily focus on transferring grasp pose representations across grippers, while overlooking strategy-level differences induced by gripper morphology~\cite{attarian2023geometry,yuan2024cross}. Incorporating gripper-aware reasoning into VLAs, where strategies span from approach trajectories to final execution, remains underexplored.

\begin{table}[t]
%\vspace{-4.5ex}
\centering
\caption{\textbf{Comparison of existing datasets for VLAs.} 
Our dataset provides a manually collected, gripper-aware data spanning multiple gripper types with gripper-specific manipulation strategies in both simulation and real-world settings.
}
% \textbf{UPGRADE THIS TABLE, IT LOOKS VERY NAIVE}Existing datasets predominantly use a single parallel jaw gripper type, while our dataset incorporates multiple gripper types to enable gripper-aware policy learning.}

\vspace{-2ex}
\label{tab:dataset_comparison}
\resizebox{\linewidth}{!}{
\begin{tabular}{l@{\hspace{8pt}}c@{\hspace{8pt}}c@{\hspace{8pt}}c@{\hspace{8pt}}c@{\hspace{8pt}}c@{\hspace{8pt}}c@{\hspace{8pt}}c@{\hspace{8pt}}c}
% \begin{tabular}{lcccccccc}
\toprule
\textbf{Dataset} & \textbf{\#Traj.} & \textbf{\makecell{\#Gripper\\ Types}} & \textbf{\#Robots} & \textbf{\makecell{Gripper-Specific\\Solution}} & \textbf{\makecell{Depth\\Image}} & \textbf{\makecell{Sub-step\\Decomposition}} & \textbf{\makecell{Data\\ Domain}} & \textbf{Collection}  \\
\midrule
Open X-Embodiment~\cite{o2024open} & 1M+ & 1 & 22 & \xmark & \xmark & \xmark &Real & Aggregation  \\
DROID~\cite{khazatsky2024droid} & 76K  & 1 & 1 & \xmark& \xmark & \xmark& Real & Human  \\
Bridge V2~\cite{walke2023bridgedata} & 60K  & 1 & 1  & \xmark& \xmark & \xmark & Real  & Human\\
RT-1~\cite{brohan2022rt} & 130K  & 1 & 1 &\xmark & \xmark & \xmark& Real & Human  \\
RH20T~\cite{fang2023rh20t} & 110K  & 1 & 7  & \xmark& \cmark  & \xmark & Real & Human \\
BridgeDataV2~\cite{fang2023rh20t} & 60.1K  & 1 & 1  & \xmark& \cmark  & \xmark & Real & \makecell{84\% Human,\\
16\% Scripted }\\
RoboMind~\cite{wu2024robomind} & 107K  & 2 & 4  & \xmark& \cmark  & \cmark & Real & Human  \\
GraspVLA~\cite{deng2025graspvla} & 1B  & 1 & 1  & \xmark & \xmark  & \xmark & Sim & Synthetic\\
Libero~\cite{liu2023libero} & 4.5K& 1 & 1  & \xmark & \xmark  & \xmark & Sim & Human\\
% Language-Table~\cite{langtable} & 440,000+ & Parallel jaw & 1 & 1 & Real \\
\midrule
\textbf{MiGA (Ours)} & \textbf{103K}  & \textbf{5} & \textbf{5}& \cmark & \cmark& \textbf{\cmark }& \textbf{Real+Sim} & \textbf{Human}  \\
\bottomrule
\end{tabular}
}
\vspace{-5ex}
\end{table}

\noindent\textbf{VLAs and Datasets.} VLAs leverage pretrained vision-language models to enable reasoning and generalization in robotic tasks. Early work like RT series~\cite{brohan2022rt,zitkovich2023rt,zhou2025rt} demonstrated this potential, while recent efforts such as OpenVLA~\cite{kimopenvla,kim2025fine} and $\pi$ series~\cite{black2410pi0,intelligence2025pi,intelligence2025pi_} have further improved scalability and architectural design. Recent work has demonstrated strong performance in grasping tasks~\cite{deng2025graspvla,zhong2025dexgraspvla,zhu2025vla}, with efforts extending to vacuum grippers~\cite{zhou2025vacuumvla} and dexterous hands~\cite{zhong2025dexgraspvla,cui2025end,li2025scalable,pan2024vision}. However, VLA models strongly rely on training data. As shown in Table~\ref{tab:dataset_comparison}, existing large-scale robot datasets predominantly feature parallel-jaw grippers~\cite{o2024open,khazatsky2024droid,walke2023bridgedata,brohan2022rt,vuong2024grasp,fang2023rh20t}, limiting VLA models' ability to learn morphology-dependent manipulation strategies~\cite{casas2024multigrippergrasp,zheng2025x}. To address this limitation, we introduce a multi-gripper dataset across five different gripper types. Unlike prior datasets, our dataset captures trajectory-level strategy variation across grippers, providing the foundation for gripper-aware policy learning.

\noindent\textbf{Soft Prompt Learning.}
Soft prompting provides a parameter-efficient alternative to full fine-tuning by optimizing continuous prompt embeddings while freezing backbone parameters~\cite{li2021prefix,liu2022late,zhou2022conditional,liu2022p}. In vision-language models (VLMs), prompt learning has shown strong few-shot adaptation performance~\cite{zhou2022learning,zhou2022conditional,gu2023systematic}, with extensions to dynamic routing~\cite{du2025mixture}, multi-modal prompts~\cite{khattak2023maple}, and knowledge preservation~\cite{yao2023visual}. Hierarchical prompt tuning~\cite{wang2022hpt} further models structured semantic associations across multiple levels via relationship-guided attention,modeling both fine-grained attributes and holistic category semantics. In robotics, prompt-based adaptation has been explored for embodiment generalization; X-VLA~\cite{zheng2025x} introduces per-embodiment soft prompts to absorb hardware variations across platforms. However, prior work focuses on robot-level diversity and overlooks end-effector morphology. We address this gap with gripper-aware soft prompt tokenization that encodes gripper taxonomy, embedding morphological priors directly into prompt space, and combine it with gripper-conditioned mixture-of-adapters~\cite{wang2022adamix,yu2024boosting} to enable structured sharing and specialization.

\vspace{-3ex}

\section{The MiGA Dataset} \label{Sec_dataset}
\vspace{-2ex}

Although several datasets have been proposed for VLA, most of them are limited to one gripper type, typically parallel-jaw grippers, and therefore overlook alternative grippers and how they influence the grasping strategy (Table~\ref{tab:dataset_comparison}). To address this limitation, we introduce MiGA, a multi-gripper-aware dataset that explicitly captures the coupling between gripper morphology and grasping strategy. MiGA has three key properties: (\textit{i}) Multi-gripper coverage: demonstrations collected with five common gripper types; (\textit{ii}) Gripper-specific task design: A broad set of tasks intentionally constructed to elicit morphology-dependent solutions, highlighting differences in contact formation, approach planning, and manipulation execution; and (\textit{iii}) Real-to-sim parity: real-world demonstrations accompanied by closely matched simulation environments that enable reproducible benchmarking across models. Beyond trajectories, MiGA provides sub-step decompositions and annotated failure demonstrations, offering rich supervision for learning gripper capabilities and challenging manipulation tasks.

\vspace{-3ex}
\subsection{Data Collection Setup}
\vspace{-1ex}

\textbf{Gripper Setup.} We employ five gripper types for data collection: (\textit{i}) parallel-jaw grippers (Franka Panda~\cite{franka2017panda}, Robotiq 2f-85~\cite{robotiq2f85}) that achieve two-finger form-closure; (\textit{ii}) three-finger grippers (Robotiq 3-Finger~\cite{robotiq3finger}) that provide multi-point form-closure; (\textit{iii}) our in-house soft two finger gripper that provides compliant adaptation; (\textit{iv}) suction gripper (Cobot Pump~\cite{cobot}, UR10 Suction Cup~\cite{ur10_suction}) that adhesion-based grasping with suction; and (\textit{v}) dexterous five fingers hand (Inspire Hand~\cite{shadowhand}) that provide high degree-of-freedom (DoF) multi-contact. These grippers differ not only in geometry but also in contact mechanics, force transmission, and controllable DoF. As a result, identical manipulation tasks require qualitatively distinct strategies across grippers, including variations in approach direction, contact region selection, and pre-grasp configuration. %For example, a suction gripper relies on surface-normal alignment and seal feasibility, whereas a parallel-jaw gripper relies on opposing contacts and stable force closure. 
With several gripper types, our dataset enables learning not only trajectory policies, but the structured relationship between grippers and grasping strategies.

\noindent\textbf{Robot Setup.}
MiGA includes demonstrations collected in both simulation and real-world settings with five gripper types. Simulation provides a reproducible testbed for algorithm development, while real-world data captures the real physical complexities. For simulation, we use NVIDIA Isaac Lab~\cite{mittal2023orbit} with Franka Panda and UR10 robots. Real-world demonstrations are collected using UFACTORY xArm7, Franka Panda, and UR5 robots. All setups use multi-view RGB-D observations from wrist-mounted cameras and third-view cameras, providing complementary end-effector and global scene perspectives. 
\vspace{-2ex}
%Camera extrinsics are calibrated per gripper configuration.

\begin{figure}[t]
    \vspace{-3ex}
    \centering
    \begin{subfigure}[b]{0.47\textwidth}            
            \includegraphics[width=\textwidth]{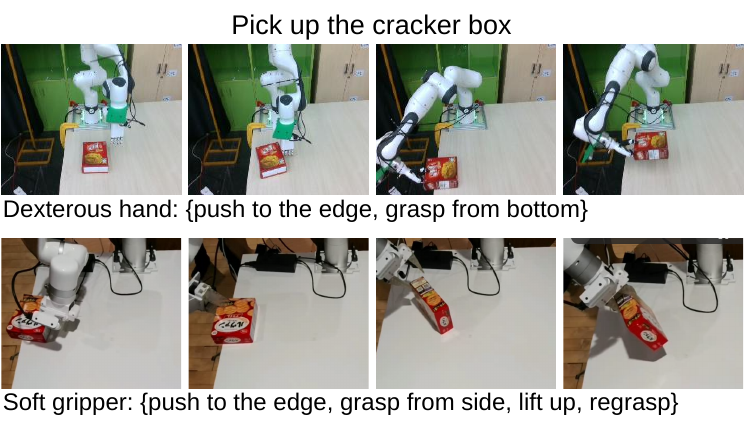}
            % \vspace{-2ex}
            \caption{Singulate}
            \label{fig:singulate}
    \end{subfigure}%
    \quad
    \begin{subfigure}[b]{0.47\textwidth}            
            \includegraphics[width=\textwidth]{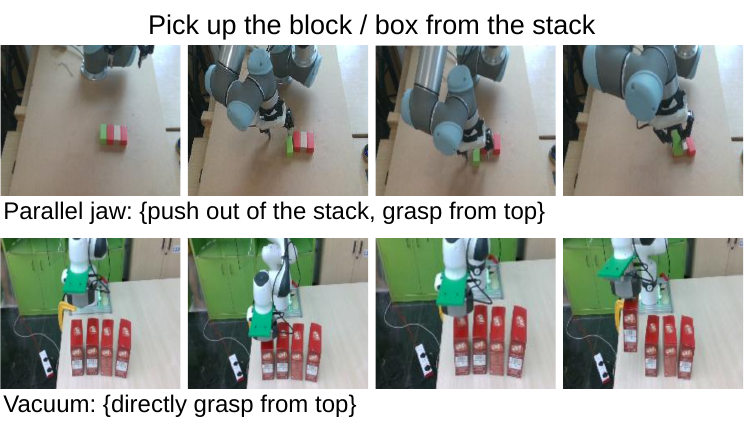}
            \caption{Stacked}
            \label{fig:stacked}
    \end{subfigure}%
    \quad
    \begin{subfigure}[b]{0.47\textwidth}            
            \includegraphics[width=\textwidth]{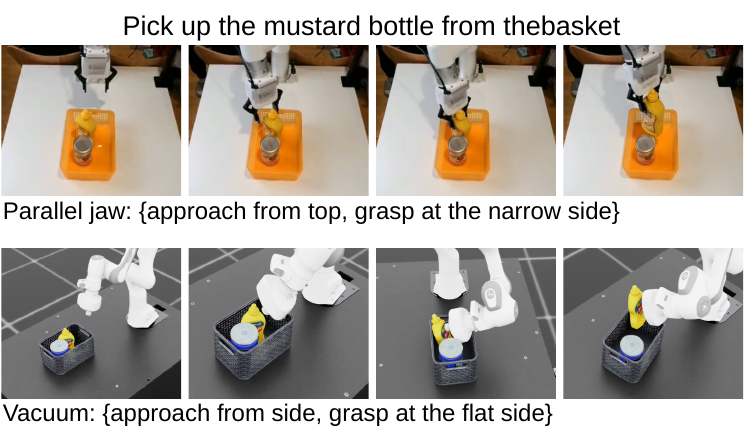}
            \caption{Constrained}
            \label{fig:constrained}
    \end{subfigure}%
    \quad
    \begin{subfigure}[b]{0.47\textwidth}            
            \includegraphics[width=\textwidth]{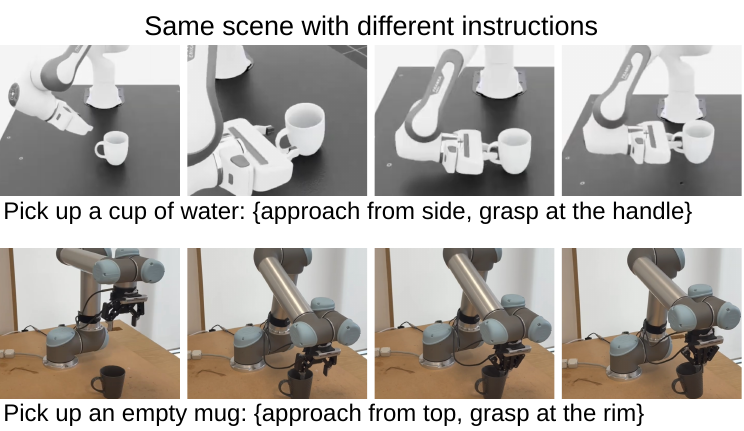}
            \caption{Semantic}
            \label{fig:semantic}
    \end{subfigure}%
    \vspace{-2ex}
    \caption{Illustration of gripper-specific strategy variations across four task categories. 
    % Sub-step (with XX gripper) {Step 1, Step 2, Step 3}
    % %(a) Singulated grasping reveals geometry-driven approach differences, (b) Stacked scenes require distinct extraction strategies, (c) Constrained spaces impose morphology-dependent collision constraints, and (d) Semantic tasks demand gripper-specific contact region selection. %\textbf{REMOVE THE BOX in each four, add as sub-figure to get (a), (b), and bigger content image}
    % %Note how identical objectives.
    }
    \label{fig:task suit}   
    \vspace{-5ex}
\end{figure}

\subsection{Task Design}
\vspace{-1ex}
To explore how gripper morphology drives grasping strategy, we design tasks in which identical objectives require different solutions across grippers. The task suite spans four scenarios: (\textit{i}) Singulated tasks vary object geometry, texture, and pose, revealing how contact feasibility depends on morphology, e.g., flat surfaces favor adhesion-based grippers while irregular geometries require compliance or multi-point contact. (\textit{ii}) Stacked scenes introduce occlusion and collision constraints that demand morphology-specific pre-grasp planning and insertion strategies. (\textit{iii}) Constrained-space tasks impose spatial limitations that amplify trade-offs between gripper size, reachability, and alignment precision. (\textit{iv}) Semantic grasping further requires reasoning about object function and task context, such as selecting safe contact regions when handling a filled container.  Our Supplementary Material provides a detailed description of these tasks. %\textbf{TODO: keep a section in Sup for this}.

%These scenarios systematically expose the coupling between embodiment and manipulation strategy, enabling evaluation of models that explicitly reason about end-effector morphology.

% \begin{figure*}[t]
%     \centering
%     \includegraphics[width=0.99\linewidth]{figures/data_distribution.pdf}
%     \vspace{-2ex}
%     \caption{\textbf{UPGRADE THIS FIGURE} GRAG dataset distribution. Comprehensive coverage of 5 morphologically diverse grippers across 4 task categories spanning 30 tasks, with balanced demonstrations for each gripper-task pair. Unlike prior single-gripper datasets, this cross-product structure enables direct supervision for learning gripper-strategy coupling across diverse manipulation scenarios.}
%     \label{dataset_distribution}
%     \vspace{-4ex}
% \end{figure*}

\begin{figure}[h]
\vspace{-3ex}
\centering
    \begin{subfigure}[b]{0.65\textwidth}            
            \includegraphics[width=\textwidth]{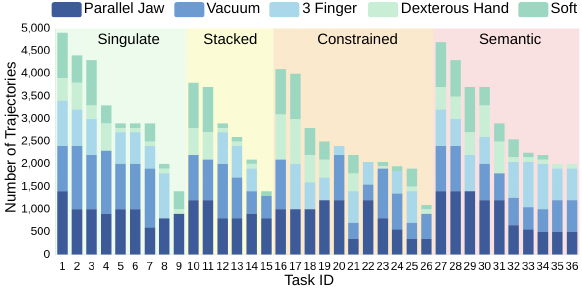}
            \caption{Distribution of tasks}
            \label{fig:SRl}
    \end{subfigure}%
    \quad
    \begin{subfigure}[b]{0.30\textwidth}
            \centering
            \includegraphics[width=\textwidth]{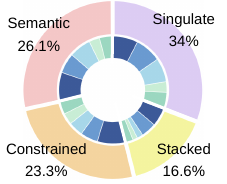}
            \vspace{0.5ex}
            \caption{Dataset composition}
            \label{fig:data distribution}
    \end{subfigure}

    \vspace{-2ex}
    \caption{MiGA dataset statistics. % (a) Demonstration coverage across 31 tasks and 5 gripper types. All tasks include at least 3 gripper types, providing multi-strategy solutions to identical objectives. (b) Balanced distribution across task categories, with all 5 gripper types represented in each category. Unlike prior datasets trained on single gripper types, this systematic structure enables direct supervision for learning the gripper-strategy coupling.}
    \label{fig:dataset_statistic}
    }
    \vspace{-5ex}
\end{figure}

\vspace{-4ex}
\subsection{MiGA Dataset Statistics}
\vspace{-2ex}
Fig.~\ref{fig:dataset_statistic} summarizes the statistics of our dataset. MiGA comprises $103,000$ demonstrations spanning $36$ tasks and $5$ distinct gripper types, with multi-view RGB-D observations, proprioceptive states, and gripper-strategy annotations. Each task includes demonstrations from at least 3 distinct gripper types, ensuring multi-strategy supervision under identical task objectives. In addition, MiGA provides natural language descriptions for each gripper-task pair. The dataset additionally includes  failure demonstrations (around $5\%$ of the dataset), enabling analysis of embodiment limitations and failure boundaries.

\vspace{-3ex}
\subsection{How will MiGA be useful to the community?}
\vspace{-1ex}
With its large-scale and multi-gripper nature, MiGA can be used in several research tasks. Here we highlight three key tasks that can be useful from our dataset: (\textit{i}) Complex grasping with trajectory-level planning: Prior work primarily targets grasp pose prediction~\cite{wu2020grasp,fang2023anygrasp,fang2020graspnet}, overlooking the grasping complexity on the whole grasping process~\cite{han2024fetchbench,murali2025graspgen}, MiGA provides full trajectories, sub-action decomposition, enabling learning of both execution policies and high-level planning. (\textit{ii}) Cross-gripper learning: Existing cross-gripper studies focus on grasp pose transfer between multi-finger grippers~\cite{khargonkar2024robotfingerprint,freiberg2025diffusion,bauer2025latent}. MiGA supports strategy comparison across different grippers, enabling morphology-dependent strategy learning beyond geometric adaptation. (\textit{iii}) VLA benchmarking: Most VLAs are trained on parallel-jaw dominant data~\cite{o2024open,brohan2022rt,wang2024scaling}, MiGA enables training VLAs with diverse grippers and provides a benchmark for evaluating cross-gripper generalization and gripper-conditioned policy learning.

\vspace{-3ex}

\section{Gripper-aware Vision Language Action Model}
\label{Sec: method}
\begin{figure*}[t]
    \vspace{-3ex}
    \centering
    \includegraphics[width=0.98\linewidth]{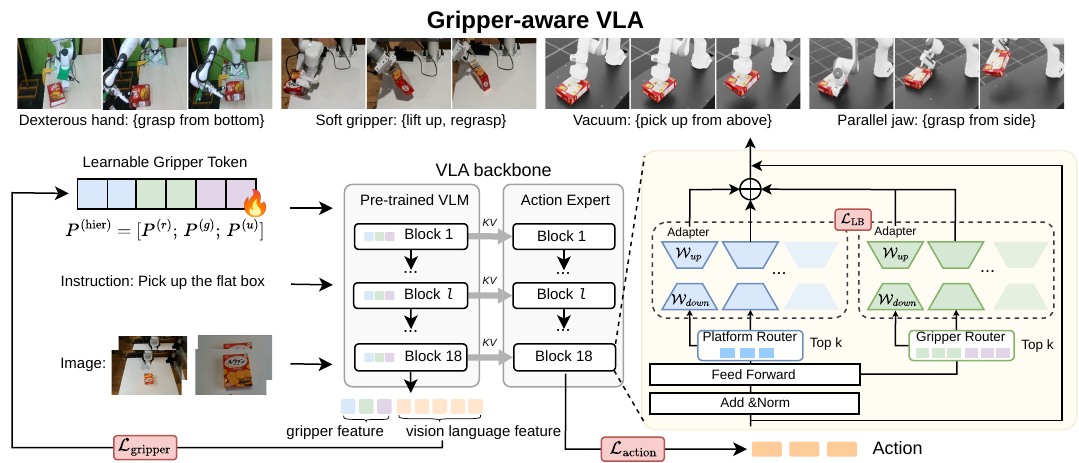}
    \caption{An overview of GVLA architecture. Our method extends the pretrained VLA backbone with two gripper-aware components. (Left) Multi-gripper tokenizer encodes gripper knowledge through three-level soft prompts: platform-specific, gripper type-specific, and instance-specific learnable tokens. (Right) A dual MoA routes action tokens through type-stratified expert pools using a cascaded domain and a gripper router, applying top-k selected adapters as residual connections to adapt to gripper-specific manipulation strategies.
    }
    \label{fig:main_method}
    \vspace{-5ex}
\end{figure*}

\vspace{-3ex}
To demonstrate the importance of gripper-aware learning in robotics and the usefulness of our MiGA dataset, we introduce GVLA, a framework that aims to include gripper information into existing VLA models. We first introduce a multi-gripper tokenizer to embed the gripper information, allowing the VLA backbone to align high-level reasoning with gripper morphology and constraints while preserving shared representations across platforms. Second, we introduce execution-level control via a dual MoA, which is routed based on gripper tokens, enabling parameter-efficient fine-tuning. GVLA is designed to be embedded into different existing VLA backbones. In practice, we choose the $\pi_{0.5}$ backbone as it shows the competitive accuracy. Fig.~\ref{fig:main_method} shows an overview of our method.

%GVLA first employs hierarchical soft prompts to condition vision-language representations for strategy reasoning, while mixture-of-adapter modulation steers action generation based on gripper features. This design enables knowledge sharing across grippers while retaining gripper-specific specialization through parameter-efficient adaptation modules, achieving effective cross-gripper transfer. 

% \subsection{Overall Framework}
% GVLA builds on the pretrained Pi0.5 architecture, which consists of a vision-language backbone for reasoning and an action expert for continuous control generation~\cite{intelligence2025pi_}. Given observation $\mathcal{O}_t$ and gripper specification tokens $(r,g,u)$ corresponding to platform, gripper type, and instance, we introduce gripper awareness at two stages of the pipeline (Fig.~\ref{fig:main_method}).

% First, we perform representation-level gripper conditioning by prepending hierarchical gripper tokens to the observation embeddings, allowing the backbone to align high-level reasoning with gripper morphology and constraints while preserving shared representations across platforms. Second, we introduce execution-level control modulation via a gripper-aware MoA inserted at the most gripper-sensitive layer of the action expert. Lightweight adapters are routed based on gripper tokens, enabling parameter-efficient specialization.

% Overall, GVLA decouples strategy reasoning from gripper-specific control adaptation, enabling structured cross-gripper transfer without full model retraining. The following subsections detail the tokenization scheme, routing mechanism, and training objectives.

% \lipsum[1]

\vspace{-2ex}
\subsection{Gripper Representation}
\label{sec:hgp}
Given observation $\mathcal{O}_t$, to condition the observation with gripper configuration. we introduce a multi-granularity representation scheme that factorises embodiment conditioning across three levels of granularity. At time $t$, the observation
$\mathcal{O}_t$ is embedded as $X \in \mathbb{R}^{n_{\text{obs}} \times d}$, where $n_{\text{obs}}$ denotes the observation token length and $d$ denotes the transformer embedding dimension.
% \noindent\textbf{Soft prompts for gripper conditioning.} 
Instead of encoding embodiment variation using fixed text prompts or general encoders such as MLPs, we employ soft prompts: learnable embeddings that are randomly initialized and optimized end-to-end through gradient-based training. Jointly optimized with the backbone under gripper prediction and action losses, these prompts learn a latent mapping $\Phi: \mathcal{H} \rightarrow \mathbb{R}^{p \times d}$ from gripper hardware configurations to a continuous prompt space, where $P^{(h)} \approx \Phi(h)$ for gripper configuration $h$. 

% \noindent\textbf{Multi-granularity factorization.}
To represent gripper characteristics, we decompose embodiment information into three independent sets of learnable embeddings at different granularity levels. Platform token $P^{(r)} \in \mathbb{R}^{p_r \times d}$, indexed by robot platform $r \in \mathcal{R}$, capture platform-specific kinematic structure and configuration. Gripper-type token $P^{(g)} \in \mathbb{R}^{p_g \times d}$, indexed by end-effector category $g \in \mathcal{G}$, encodes mechanism-specific manipulation priors shared across all grippers of the same type. Instance-level token $P^{(u)} \in \mathbb{R}^{p_u \times d}$, indexed by gripper instance $u \in \mathcal{U}$, models fine-grained characteristics unique to individual grippers. The prompts are concatenated using a predefined three-level order:
\begin{equation}
P^{(\text{h})} = [P^{(r)};\, P^{(g)};\, P^{(u)}]
\in \mathbb{R}^{(p_r + p_g + p_u) \times d},
\end{equation}
and prepended to the embedded observation to form the conditioned input:
\begin{equation}
\tilde{X}^{(r,g,u)} = [P^{(\text{h})};\, X]
\in \mathbb{R}^{(p_r + p_g + p_u + n_{\text{obs}}) \times d}.
\end{equation}

This multi-granularity factorization encodes gripper information at three levels of specificity, capturing shared knowledge across different gripper configurations at each level, which facilitates efficient adaptation to new gripper instances.

% disentangling shared structure across platforms and grippers from task-specific variation. As a result, knowledge learned at different level tokens can potentially be more useful for being reused, which facilitates adaptation when new gripper instances or platform–gripper combinations are introduced at test time.

\begin{wrapfigure}{r}{0.36\textwidth}
% \vspace{-3ex}
    \centering
    \includegraphics[width=0.37\textwidth]{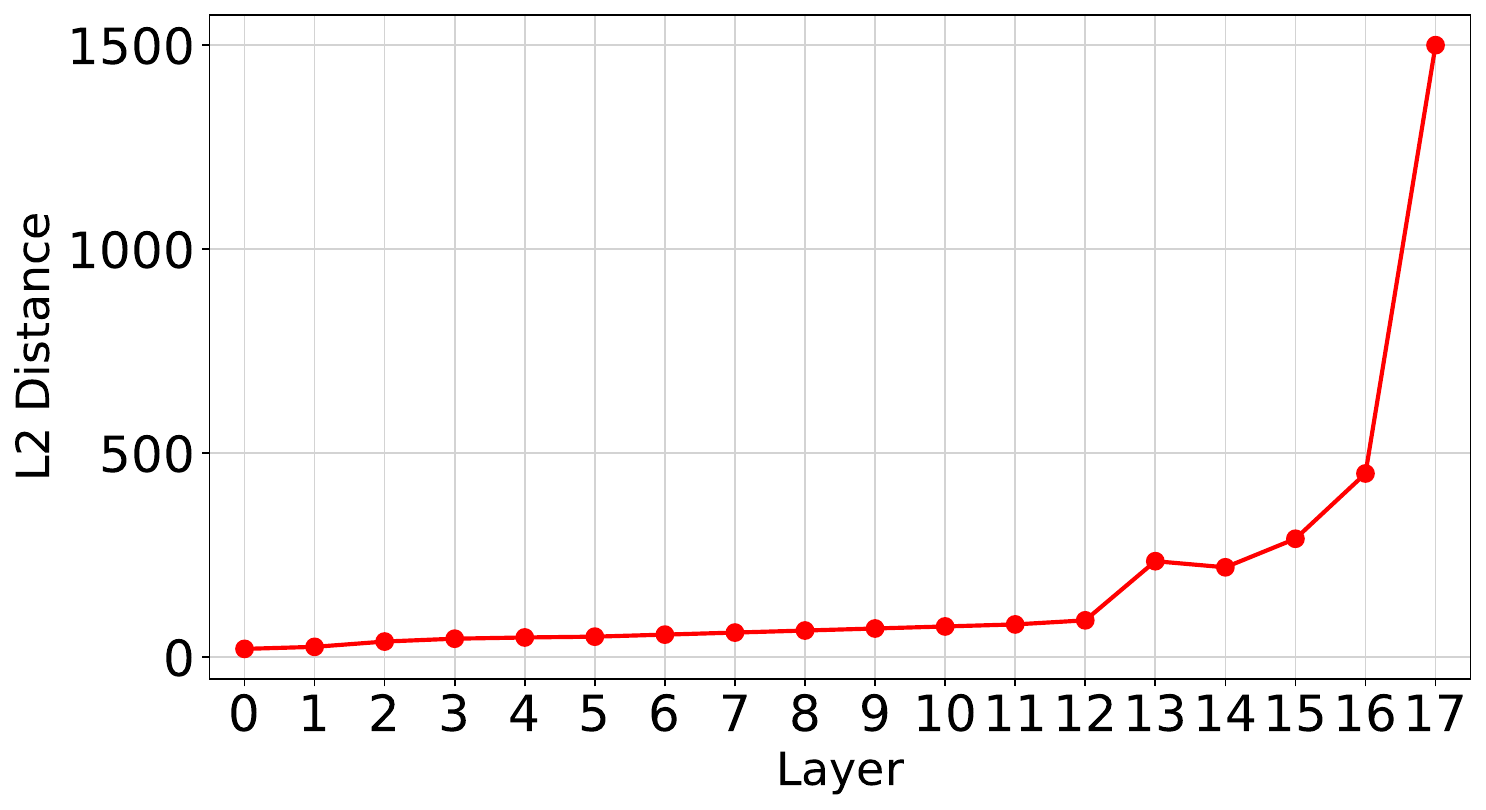}
    \vspace{-5ex}
    \caption{Layer probing analysis.}
    \label{fig:layer analysis}
    \vspace{-5ex}
\end{wrapfigure}

\subsection{Gripper-Aware Mixture of Adapters}
\label{sec:gripper_routing}
While our multi-gripper tokenizer provides high-level conditioning, effective execution requires modulating action generation. We introduce a dual MoA mechanism that decomposes gripper-aware adaptation into platform-specific and gripper-specific modulation. %We insert dual MoA modules at selected transformer layers $\mathcal{L} = \l_K\}$ within the action expert.

Each MoA computes a gating function over the pooled conditioning tokens:
\begin{equation}
    G(P) = \text{Softmax}(\text{TopK}(\text{MLP}(\text{mean}(P)))).
\end{equation}
We instantiate two parallel routers: (\textit{i}) a platform-aware gate $G^{(p)}$ driven 
by $\texttt{mean}(P^{(r)})$ and (\textit{ii}) a gripper-aware gate $G^{(g)}$ driven by 
$\texttt{mean}([P^{(g)};\, P^{(u)}])$, each returning normalised weights and 
selected expert indices.

Each expert implements a bottleneck transformation:
\begin{equation}
    \mathcal{A} = \mathcal{W}^{\text{up}}(\text{GeLU}(\mathcal{W}^{\text{down}}(\mathbf{x}))), 
\end{equation}
where $\mathcal{W}^{\text{down}} \in \mathbb{R}^{d \times d_b}$, $\mathcal{W}^{\text{up}} \in \mathbb{R}^{d_b \times d}$ 
with $d_b \ll d$. Given layer activation $\mathbf{x} \in \mathbb{R}^{B \times S \times d}$, 
the final output combines both platform and gripper adaptations additively:
\begin{equation}
    \mathbf{x} \leftarrow \mathbf{x} + \sum_{i=1}^{k} G^{(p)}_i \cdot \mathcal{A}^{(p)}_i + \sum_{j=1}^{k} G^{(g)}_j \cdot \mathcal{A}^{(g)}_j.
\end{equation}

%mathbf - replace

\noindent\textbf{Where to Insert MoA?} 
To insert MoA into the VLA backbone, we analyze how gripper features influence hidden representations by measuring gripper type sensitivity of the action expert layers in the VLA backbone. This is defined as:
\begin{equation}
S_{\text{type}} = \frac{1}{|\mathcal{G}|^2} 
\sum_{\substack{g_1, g_2 \in \mathcal{G} \\ g_1 \neq g_2}} 
\left\| \text{mean}_{u \in \mathcal{U}_{g_1}} h_l^u 
- \text{mean}_{u \in \mathcal{U}_{g_2}} h_l^u \right\|_2,
\end{equation}
where $h_l^u$ denotes the action hidden representation at layer $l$ for gripper $u$. Fig~\ref{fig:layer analysis} shows that the type sensitivity remains low in several early layers but increases sharply in the final layer. We therefore insert MoA at the final layer to directly modulate action generation via multi-gripper-aware conditioned routing. More studies validating this design are provided in our Supplementary Material.
% , indicating that gripper-specific execution dynamics emerge primarily in the deepest representations
% \subsubsection{Gripper Generalization.}
% This design supports adaptation to novel grippers at test time. A previously unseen gripper can be assigned to a known type category and initialized with learnable identity tokens. Through gradient-based optimization of these tokens on a small calibration dataset, the model learns appropriate routing patterns within the type's expert pool, enabling rapid adaptation without retraining the entire network.

\subsection{Fine-tuning Objective}
\label{sec:training}
To jointly learn gripper-aware action generation and gripper-consistent representations, we optimize GVLA with the following objective:
\begin{equation}
\mathcal{L} = \mathcal{L}_{\text{action}} + \lambda_{\text{gripper}} \mathcal{L}_{\text{gripper}} + \lambda_{\text{LB}} \mathcal{L}_{\text{LB}},
\end{equation}
\noindent\textbf{Action Loss.}
Following~\cite{black2410pi0}, we supervise action tokens using a conditional 
flow matching loss~\cite{lipman2022flow,liu2022rectified}:
\begin{equation}
  \mathcal{L}_{\text{action}} = \mathbb{E}\left\| v_\theta(A_t^\tau, \mathcal{O}_t) - u(A_t^\tau | A_t) \right\|^2,   
\end{equation}
where the network learns to predict the vector field $u(A_t^\tau | A_t) = \boldsymbol{\epsilon} - A_t$ that transports noisy actions $A_t^\tau = \tau A_t + (1-\tau)\boldsymbol{\epsilon}$ back to clean actions $A_t$, with noise $\boldsymbol{\epsilon} \sim \mathcal{N}(0, \text{I})$ and flow timestep $\tau \in [0, 1]$.

\noindent\textbf{Gripper Prediction Loss.}  
To encourage gripper-discriminative features in shared vision-language embeddings, we add an auxiliary classification loss. After gripper-conditioned encoding, the pooled visual observation representation $\mathcal{O}_t$ are used to predict the gripper type:
\begin{equation}
\mathcal{L}_{\text{gripper}} = -\log \frac{\exp\!\left(f_\phi^{(y_g)}(\mathcal{O}_t)\right)}{\sum_{k=1}^{|\mathcal{G}|} \exp\!\left(f_\phi^{(k)}(\mathcal{O}_t)\right)}
\end{equation}
where $f_\phi$ is a linear classifier and $y_g \in \mathcal{G}$ the ground-truth label. This loss integrates gripper-relevant information into the shared embeddings, supporting gripper-sensitive reasoning and stabilizing downstream routing.

\noindent\textbf{Load Balance Loss.}
To prevent router collapse and ensure uniform adapter utilization across the type-stratified pools, we introduce a load balance regularization. For each adapter pool, we measure the variance of adapter usage distribution and penalize imbalanced activation patterns:
\begin{equation}
\mathcal{L}_{\text{LB}} = \frac{1}{|\mathcal{R}|} \sum_{p \in \mathcal{R}} \text{Var}(\alpha_p) + \frac{1}{|\mathcal{G}|} \sum_{g \in \mathcal{G}} \text{Var}(\alpha_g),
\end{equation}
where $\alpha_p = [\alpha_p^1, \ldots, \alpha_p^{A_p}]$ denotes the mean activation frequency of the $\mathcal{A}_p$ adapters in platform pool $p$, and $\alpha_g = [\alpha_g^1, \ldots, \alpha_g^{A_g}]$ represents the usage distribution for gripper type pool $g$. This objective encourages the routers to distribute load evenly within each pool, preventing underutilization of specialized adapters while maintaining the hierarchical routing structure.

% where we set $\lambda_{\text{gripper}} = 0.1$ and $\lambda_{\text{LB}} = 0.01$.

% \subsection{Inference}
% \label{sec:inference}

% At inference, computational efficiency is achieved through caching. The prefix (including hierarchical soft prompts) is processed once and cached. FiLM parameters $(\gamma, \beta)$ are computed from cached Layer-17 features. Actions are generated through iterative denoising:
% \begin{equation}
% \mathbf{a}_{t-\Delta t} = \mathbf{a}_t + \Delta t \cdot \mathbf{v}_\theta(\mathbf{a}_t, t, \mathcal{O}, g),
% \end{equation}
% starting from $\mathbf{a}_1 = \boldsymbol{\epsilon}$ with $T=10$ Euler integration steps.

\vspace{-2ex}
\section{Experiments}
\label{sec:experiments}
\vspace{-2ex}
We conduct experiments to investigate: (\textit{i}) Overall Performance: How does gripper-aware conditioning improve manipulation success compared to baselines across task categories? (\textit{ii}) Gripper-Aware Transfer: Does our model enable efficient zero-shot generalization or few-shot adaptation? (\textit{iii}) Ablation Study and Robot Validation: How do our multi-gripper tokenizer and MoA contribute to performance, and how does GVLA perform in real-robot experiments? 

\label{subsec:exp_setup}
% \noindent\textbf{Datasets.}

\noindent\textbf{Baselines.}
To validate the effectiveness of our method, we benchmark representative models from both traditional and VLA-based methods: For two-stage open-loop grasping, we include AnyGrasp~\cite{fang2023anygrasp}, a state-of-the-art grasp detector trained on large-scale data, and GraspMAS~\cite{nguyen2025graspmas}, a recent multi-agent approach for language-driven grasp detection.
For VLA-based methods, we include GraspVLA~\cite{deng2025graspvla}, which leverages large-scale grasping data to enable zero-shot generalization across diverse scenes, and OpenVLA-OFT~\cite{kim2025fine}, an optimized variant of OpenVLA~\cite{kimopenvla}. We also implement two variants based on the $\pi_0$ architecture~\cite{black2410pi0}: the original $\pi_0$ and the improved $\pi_{0.5}$, and evaluate their performance both as vanilla baselines and with our proposed gripper-aware extensions.

\noindent\textbf{Evaluation Metrics.} We use five metrics to evaluate the results: (\textit{i}) Success Rate (SR) is defined as $\text{SR} = \frac{1}{N} \sum_{i=1}^{N} S_i$; (\textit{ii}) The Prediction Error (PE) measures the average deviation between predicted and ground-truth actions over the action horizon $H$: $\text{PE} = \frac{1}{H}\sum_{h=1}^{H} |a_{\text{pred}}^{(h)} - a_{\text{gt}}^{(h)}|$; (\textit{iii}) To measure behavioral specialization, we use the Counterfactual Action Prediction Divergence (CAPD)~\cite{veitch2021counterfactual}, which quantifies how much predicted actions change when only the gripper identity is modified while all other inputs remain fixed.
% \begin{equation}
% \text{CAPD} =
% \frac{2}{K(K-1)}
% \sum_{i<j}
% \frac{
% \left\| a(g_i) - a(g_j) \right\|_2
% }{
% \left\| a_{\text{mean}} \right\|_2
% },
% \end{equation}
% where $K$ is the number of gripper types, $a(g_i)$ is the predicted action for gripper $g_i$, and $a_{\text{mean}}$ is the mean across 
% grippers. 
Higher CAPD indicates gripper-specific behavior; (\textit{iv})
To understand how gripper information is encoded throughout the network, we measure Linear Probe Accuracy (LPA)~\cite{alain2016understanding}, which trains a logistic regression classifier $\phi^{(l)}$ on hidden states at each transformer layer $l$; trained on in-distribution data
and evaluated on unseen gripper (see supplementary for full definitions); and (\textit{v}) To quantify the contribution of the gripper token, we further
introduce the Gripper Contribution Score (GCS): 
$ \text{GCS} =
    \text{LPA}^{(l)}\!\left(\tilde{g} = g_i\right) -
    \text{LPA}^{(l)}\!\left(\tilde{g} = g_\text{fixed}\right)
$. A large GCS indicates that the soft prompt contributes to gripper discrimination
beyond visual features alone.

\vspace{-3ex}
\subsection{Main Result} 
\vspace{-2ex}

\begin{table}[t]
  \centering
  \caption{\textbf{Baseline comparison across task categories.}}
  \vspace{-2ex}
  \scalebox{0.9}{
  \begin{tabular}{lccccc}
    \toprule
    Method & Flat & Stacked & Constrained & Semantic & Avg.(\%) \\
    \midrule
    AnyGrasp~\cite{fang2023anygrasp}        & 0.00  & 0.00  & 46.00 & 0.00  & 11.50 \\
    GraspMAS~\cite{nguyen2025graspmas}      & 0.00  & 0.00  & 40.00 & 8.00  & 12.00 \\
    \midrule
    GraspVLA~\cite{deng2025graspvla}        & 1.50  & 0.00  & 30.00 & 0.00  & 7.88  \\
    OpenVLA-OFT~\cite{kim2025fine}                     & 41.00 & 52.50 & 24.00 & 50.00 & 41.88 \\
    $\pi_{0}$~\cite{black2410pi0}           & 30.00 & 21.50 & 36.00 & 65.00 & 38.13 \\
    $\pi_{0.5}$~\cite{intelligence2025pi_}   & \underline{52.50} & \underline{71.00} & \underline{57.50} & 52.50 & \underline{58.38} \\
    \midrule
    GVLA ($\pi_0$ backbone) (Ours)                   & 27.50 & 45.00 & 50.00 & \underline{70.00} & 48.13 \\
    \rowcolor{gray!15}
    \textbf{GVLA} ($\pi_{0.5}$ backbone) (Ours)               & \textbf{53.00} & \textbf{76.00} & \textbf{62.50} & \textbf{72.50} & \textbf{66.00} \\
    \bottomrule
  \end{tabular}
  }
  \label{tab:baseline}
  \vspace{-4ex}
\end{table}

Table~\ref{tab:baseline} compares our GVLA, fine-tuned on our MiGA dataset from $\pi_0$ and $\pi_{0.5}$ backbones against different baselines across four task categories across different gripper types in simulation. This table shows that traditional two-stage methods (AnyGrasp~\cite{fang2023anygrasp}, GraspMAS~\cite{nguyen2025graspmas}) collapse entirely on flat and stacked scenes, exposing their limited capacity to generalize beyond simple grasp configurations. Despite large-scale pre-training, GraspVLA~\cite{deng2025graspvla} encounters similar degradation, highlighting the importance of our dataset design, which encodes diverse gripper-dependent grasping strategies instead of relying on a single downward grasping prior.
Among VLA-based baselines, $\pi_{0.5}$ achieves the highest average performance (58.38\%), yet remains limited. Leveraging this backbone, GVLA improves performance by 7.62\% and surpasses all competing approaches
% . Importantly, GVLA maintains relatively uniform performance across gripper types
, suggesting that gripper-aware conditioning enables the policy to capture gripper-specific manipulation strategies in scenarios where the robot behaviors differ substantially.

\begin{table}[h]
\vspace{-6ex}
  \centering
  % \begin{minipage}[b]{0.27\textwidth}  % Changed [t] to [b]
  %   \centering
  %   \includegraphics[width=\textwidth]{example-image-a}
  %   \captionof{figure}{Neuron Activation Heatmap}
  %   \label{fig:neuron_heatmap}
  % \end{minipage}%
  % \hfill
  \begin{minipage}[t]{0.48\textwidth}  % Changed [t] to [b]
    \centering
    \includegraphics[width=\textwidth]{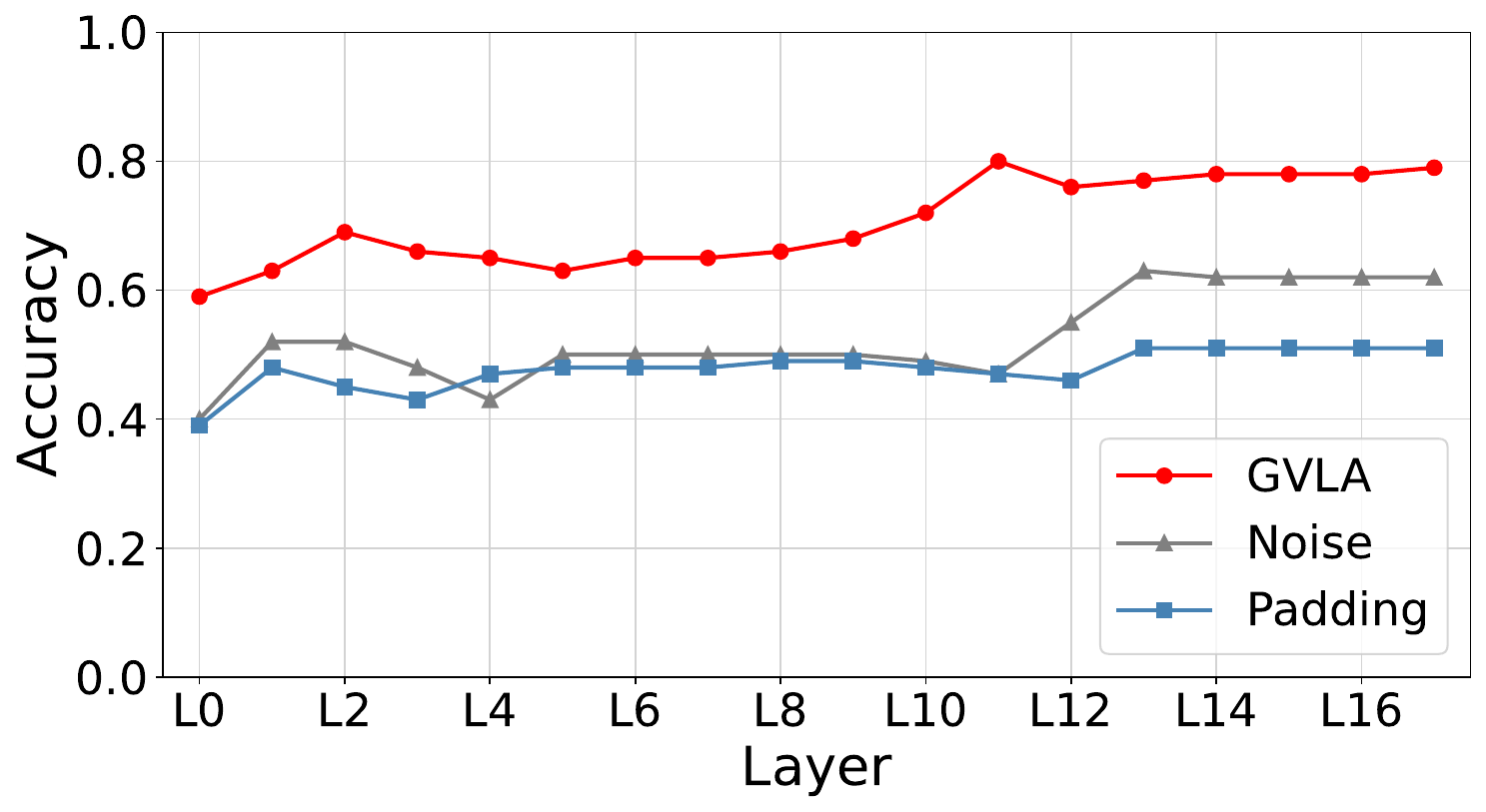}
    \captionof{figure}{Linear accuracy result.}
    \label{fig:linear_probe}
  \end{minipage}%
  \hfill
  \begin{minipage}[b]{0.48\textwidth}  % Changed [t] to [b]
    \centering
    % \vspace{-9ex}
    \caption{Tokenizer comparison.}
    \vspace{-2ex}
\scalebox{1.0}{%
    \begin{tabular}[b]{lccc}
      \toprule
      Method & PE $\downarrow$ & CAPD $\uparrow$ & GCS $\uparrow$ \\
      \midrule
      % w/o GP&0.037 & &0.0\\
      % Language & 0.38\\
      MLP~\cite{rumelhart1985learning}  & 0.053& 0.82& 0.014\\
      VQ-VAE~\cite{van2017neural} & 0.051&0.70&0.002\\
      % LP &0.053&0.64 &0.421 \\
        LP~\cite{li2024llava} &0.053&0.64 & 0.225 \\
      \midrule
          \rowcolor{gray!15}
      GVLA (Ours) &\textbf{0.032}&\textbf{1.34} & \textbf{0.249}\\
      \bottomrule
    \end{tabular}
}
    \label{tab:tokenizer}
  \end{minipage}
  \vspace{-6ex}
\end{table}

\subsection{Gripper-Aware Analysis}
\vspace{-1ex}
\noindent\textbf{Does Gripper Conditioning Steer Specialized Pathways?}
%To verify that GVLA learns genuine gripper-aware representations rather than exploiting superficial token patterns or inferring embodiment from visual cues, we conduct two complementary analyses: (1) layer-wise linear probing to evaluate representation separability, and (2) behavioral specialization metrics to quantify embodiment-conditioned policy divergence.
Fig.~\ref{fig:linear_probe} presents linear probe accuracy across network depth for three conditioning strategies. The ``Noise'' token baseline replaces gripper tokens with noise (hence, in this case, the backbone only relies on the visual and language tokens for learning) to test whether the input tokens suffice for effectively learning the tasks.
The ``Padding'' token baseline uses a fixed, gripper-agnostic token to test whether improvements arise from explicit conditioning rather than implicit visual inference. Both baselines remain low in accuracy, demonstrating that neither arbitrary tokens nor visual cues alone provide sufficient embodiment information. In contrast, our GVLA with multi-gripper tokenizer exhibits consistently elevated accuracy, rising from 59\% at layer L0 to 80\% at L12, and maintaining high separability in late layers. This persistent gripper-discriminative signal indicates that our multi-gripper tokenization enables the model to encode and propagate gripper-specific features throughout the network through explicit conditioning, rather than just relying on incidental visual patterns or treating the gripper information as auxiliary metadata.
Table~\ref{tab:tokenizer} further supports this observation: GVLA achieves the lowest prediction error and the highest behavioral divergence, indicative of learning effective gripper-specific behavior. The combination of low PE and high CAPD demonstrates that our method improves both predictive accuracy and policy specialization, enabling GVLA to discover manipulation strategies aligned with each gripper’s affordances.

\begin{figure}[t]
\vspace{0ex}
\centering
    \begin{subfigure}[b]{0.45\textwidth}            
            \includegraphics[width=\textwidth]{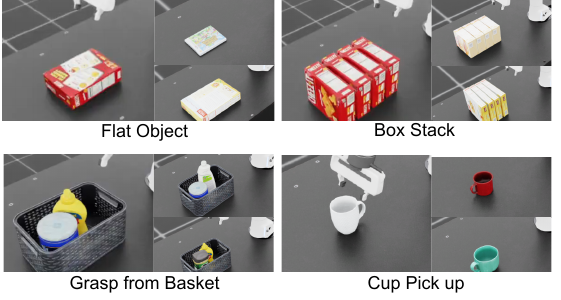}
            \caption{Cross-object generalization setup.}
            \label{fig:exp task}
    \end{subfigure}%
    \quad
    \begin{subfigure}[b]{0.45\textwidth}
            \centering
            \includegraphics[width=\textwidth]{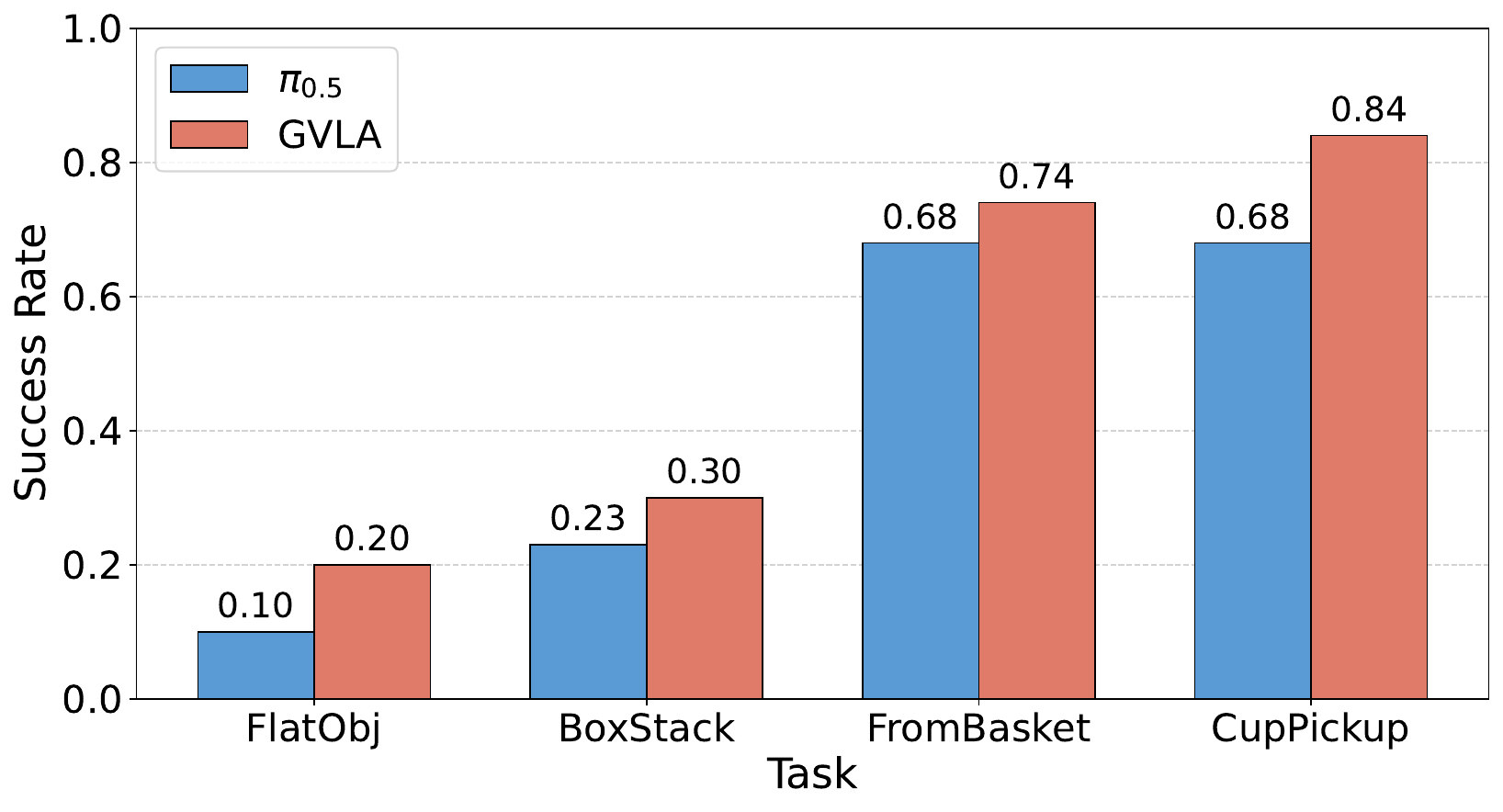}
            \caption{Cross-object generalization success rate.}
            \label{fig:generalization}
    \end{subfigure}
    \vspace{-2ex}
    \caption{Cross-object generalization results.}\label{fig:task generalization}
    \vspace{-6ex}
\end{figure}

\begin{figure}[h]
\vspace{-5ex}
    \begin{subfigure}[b]{0.31\textwidth}            
            \includegraphics[width=\textwidth]{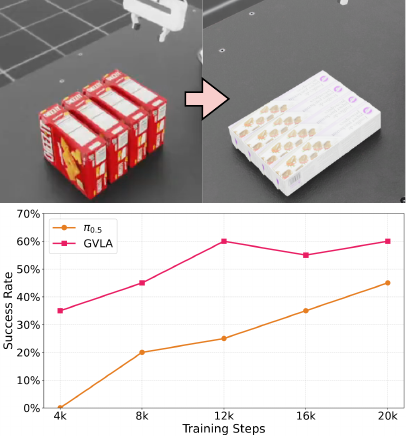}
            \caption{Task adaptation. }
            \label{fig:task adaptation}
    \end{subfigure}%
    \quad
    \begin{subfigure}[b]{0.31\textwidth}
            \centering
            \includegraphics[width=\textwidth]{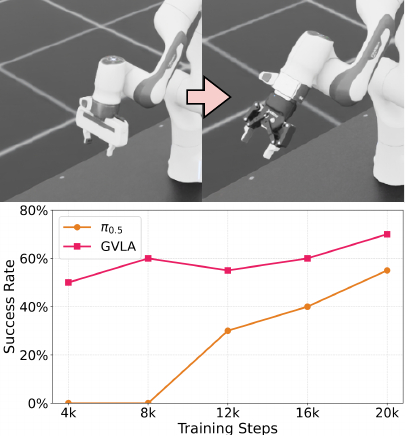}
            \caption{Gripper adaptation}
            \label{fig:gripper adaptation}
    \end{subfigure}
        \quad
    \begin{subfigure}[b]{0.31\textwidth}
            \centering
            \includegraphics[width=\textwidth]{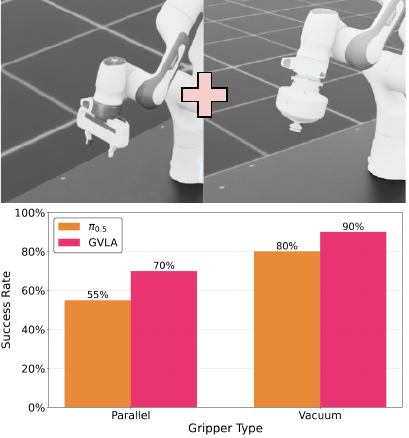}
            \caption{Mix data adaptation}
            \label{fig:mix data}
    \end{subfigure}
    \vspace{-2ex}
    \caption{Adaptation results.}\label{fig:task adaptation}
    \vspace{-5ex}
\end{figure}
\noindent\textbf{Cross-object Generalization.}
To evaluate zero-shot generalization, we construct task variants by introducing new objects that share similar functional attributes with the training set and evaluate performance. This setting measures the model's ability to generalize to new grasping scenarios without any task-specific fine-tuning. Fig.~\ref{fig:exp task} shows the setup of this experiment. As reported in Fig.~\ref{fig:generalization}, GVLA maintains strong performance across unseen tasks, demonstrating robust zero-shot transfer capability.

\noindent\textbf{Gripper-Aware Adaptation.}
To evaluate whether our model can effectively disentangle gripper-specific representations and enable knowledge transfer across grippers, we conduct few-shot adaptation experiments under three settings: 
(\textit{i}) adapting to a new task with the same gripper, 
(\textit{ii}) adapting to a new task with a previously unseen gripper, and 
(\textit{iii}) adapting to a new task using a mixture of data from two different grippers, which are jointly used for training. Our Supplementary Material provides the detailed experimental setup. Fig.~\ref{fig:task adaptation} shows that our model exhibits faster adaptation on new tasks involving the same type of gripper, as these tasks share a common adapter and underlying knowledge representation. When trained with mixed-gripper data, the model achieves further performance improvements, as jointly learning from multiple grippers reduces ambiguity and confusion between different gripper types.

\begin{wraptable}{r}{0.36\textwidth}
  \centering
  \vspace{-7ex}
  \caption{Component analysis.}
  \label{tab:component_analysis}
  \scalebox{0.72}{
    \begin{tabular}{clcc}
      \toprule
      & Configuration & PE$\downarrow$ & Adapt.$\uparrow$ \\
      \midrule
    \multirow{3}{*}{\shortstack[c]{Gripper\\Repr.}}
&w/o $P^{(g)}$                    & 0.037 & 0.86 \\
&w/o $P^{(p)}$                    & 0.035 & 0.54 \\
&w/o $P^{(u)}$                    & 0.032 & 0.90 \\
      \midrule
      \multirow{3}{*}{\shortstack[l]{Dual\\MoA}}
&w/o MoA                          & 0.037 & 0.52 \\
&w/o $\text{MoA}^{(g)}$           & 0.033 & 0.56 \\
&w/o $\text{MoA}^{(p)}$           & 0.034 & 0.54 \\
      \midrule
      \rowcolor{gray!15}
      & GVLA                      & \textbf{0.032} & \textbf{0.92} \\
      \bottomrule
    \end{tabular}
  }
  \vspace{-6ex}
\end{wraptable}

% \begin{wraptable}{r}{0.42\textwidth}
%   \centering
%   \vspace{-3ex}
%   \caption{Component analysis.}
%   \label{tab:ablation}
%   \scalebox{0.72}{
%     \begin{tabular}{clcc}
%       \toprule
%       & Configuration & PE$\downarrow$ & Adapt.$\uparrow$ \\
%       \midrule
%     \multirow{3}{*}{\shortstack[c]{Gripper\\Repr.}}
%       & $\pi_{0.5}$ (baseline)                     & 0.037 & 0.86 \\
%         & \quad + $P^{(g)}$                         & 0.035 & 0.54 \\
%         & \quad + $P^{(g)}$ + $P^{(r)}$             & 0.032 & 0.90 \\
%       \midrule
%       \multirow{3}{*}{\shortstack[l]{Dual\\MoA}}
%         & $P^{(h)}$ only                            & 0.037 & 0.52 \\
%         & \quad + $\text{MoA}^{(g)}$                & 0.033 & 0.56 \\
%         & \quad + $\text{MoA}^{(p)}$                & 0.034 & 0.54 \\
%       \midrule
%       \rowcolor{gray!15}
%       & \textbf{GVLA (full)}                        & \textbf{0.032} & \textbf{0.92} \\
%       \bottomrule
%     \end{tabular}
%   }
%   \vspace{-4ex}
% \end{wraptable}

\noindent\textbf{Component Analysis.} Table~\ref{tab:component_analysis} evaluates the contribution of each component in GVLA. We report prediction error (PE) on the pre-trained MiGA dataset and adaptation success rate (Adapt.) when transferring to an unseen gripper (Robotiq 2F-85) after 20K fine-tuning steps.

% The result shows that removing the gripper-type token $P^{(g)}$ degrades both PE and adaptation, indicating that type-level priors contribute to both prediction quality and adaptation. In contrast, removing the platform token $P_{(p)}$ sharply reduces adaptation, indicating that platform information is not critical for in-distribution fitting but is essential for disentangling embodiment-dependent dynamics during transfer. Removing the instance token $P_{(u)}$ has minimal impact, suggesting it mainly refines specialization within a fixed embodiment rather than enabling structural generalization.

The result shows that removing the gripper-type token $P^{(g)}$ degrades both PE and adaptation, indicating that type-level priors contribute to both prediction quality and adaptation. The platform token $P^{(p)}$ is critical for cross-embodiment transfer: without it, adaptation drops sharply despite comparable PE, highlighting the necessity of platform disentanglement. In contrast, removing the instance token $P^{(u)}$ only slightly lowers adaptation, suggesting it mainly provides fine-grained specialization for novel gripper instances.
Ablating the dual MoA restores PE to baseline and severely reduces adaptation, demonstrating that representation-level prompting alone is insufficient without computation-level modulation. Removing either $\text{MoA}^{(g)}$ or $\text{MoA}^{(p)}$ yields similar degradation, implying complementary routing roles. The full GVLA achieves the highest PE and adaptation score, supporting that disentangled prompting and dual routing jointly enable robust cross-gripper learning.

\begin{figure}[t]
\centering
    \begin{subfigure}[b]{0.6\textwidth}            
        \includegraphics[width=\linewidth, height=0.33\linewidth]{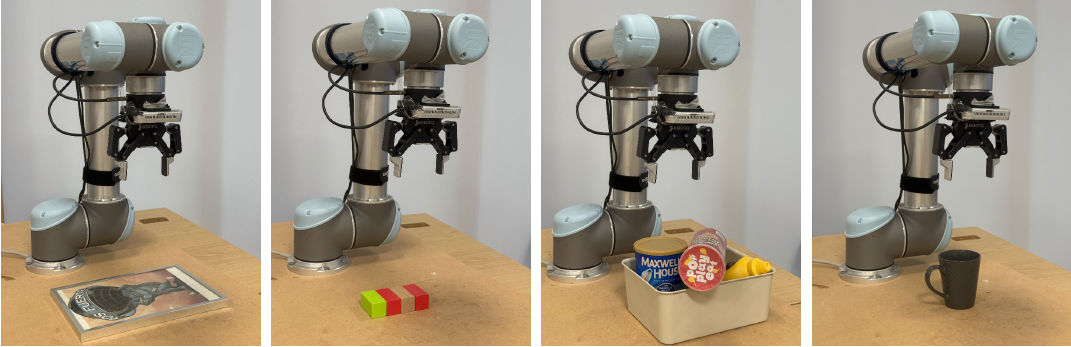}
        % \vspace{-3ex}
        \caption{Real robot experiment setup.}
        \label{fig:real-word}    
    \end{subfigure}%
    \quad
    \begin{subfigure}[b]{0.34\textwidth}
            \centering
            \includegraphics[width=\textwidth, height=0.58\linewidth]{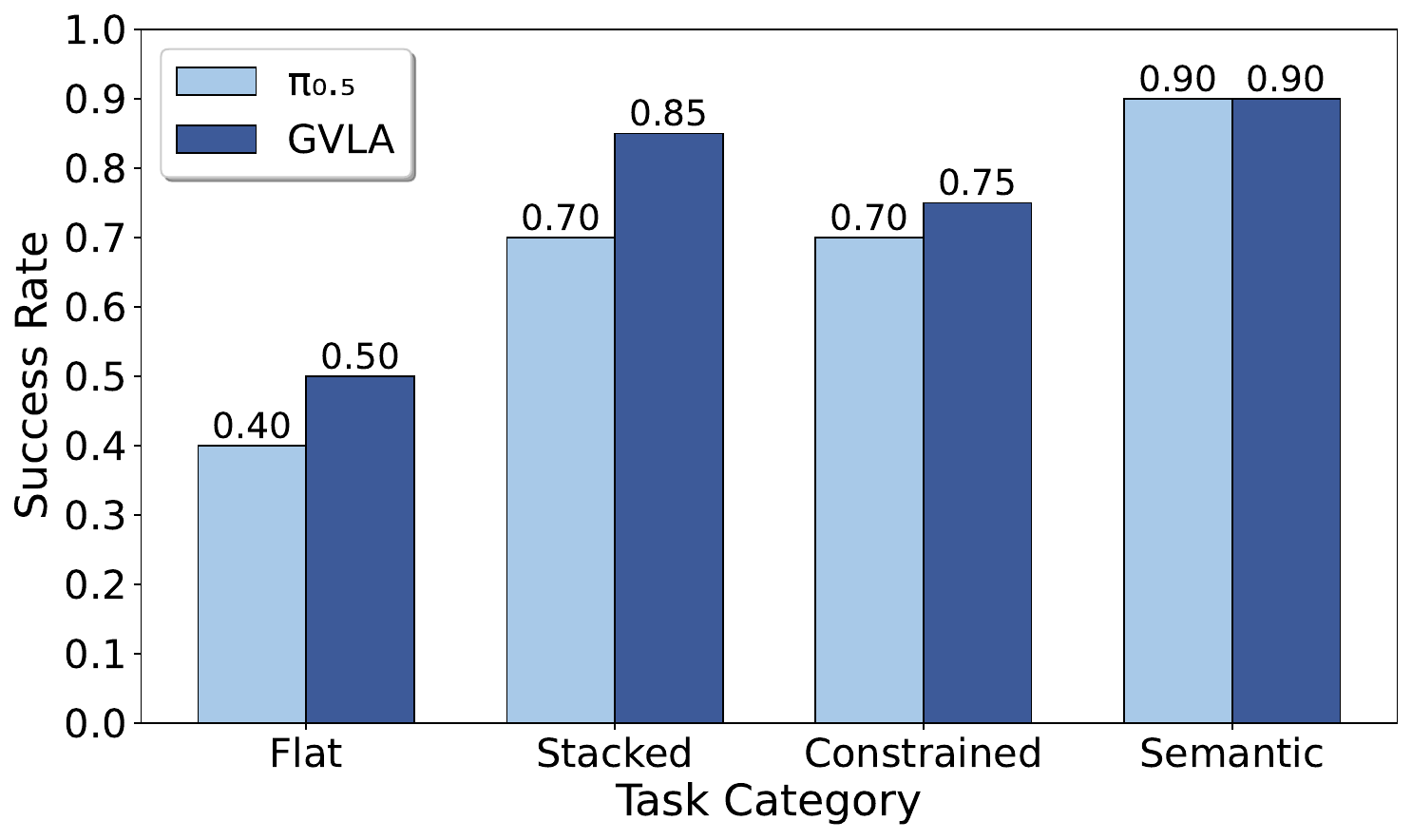}
            \caption{Real robot experiment results.}
            \label{fig:rw generalization}
    \end{subfigure}
    \vspace{-1ex}
    \caption{Real-world robotic experiment setups and results.}\label{fig:task generalization}
    \vspace{-5ex}
\end{figure}
\subsection{Real-world Robotic Validation} 
% \noindent\textbf{Real-world Experiment.} 
\noindent\textbf{Robot Results.} 
To validate whether gripper-aware conditioning transfers effectively to real-world settings, we conduct a targeted evaluation on cross-domain task generalization: we evaluate our model on a real UR5 arm equipped with a Robotiq 2f-85 gripper, adapting to tasks unseen during training using only 10 demonstrations and 20k fine-tuning steps. Fig.~\ref{fig:real-word} shows the setup of our robotic experiment. 
% This setting isolates the contribution of gripper-aware representations to data-efficient sim-to-real transfer under a genuine embodiment shift. 
As shown in Fig.~\ref{fig:rw generalization}, our GVLA outperforms the $\pi_{0.5}$ baseline across all tasks, demonstrating that gripper-aware conditioning facilitates embodiment-specific knowledge transfer, enabling efficient few-shot adaptation to novel tasks across unseen platform-gripper configurations.

% \begin{table}[t]
%   \centering
%   \caption{\textbf{real-world experiment.}}
%   \vspace{-3ex}
%   \scalebox{0.7}{
%   \begin{tabular}{lcccccccccccc}
%     \toprule
%      & \multicolumn{2}{c}{Flat Object}
%      & \multicolumn{2}{c}{Stacked Objects}
%      & \multicolumn{2}{c}{Constrained Space}
%      & \multicolumn{2}{c}{Semantic Grasping }\\
%     \cmidrule(r{6pt}){2-3}
%     \cmidrule(l{6pt}r{6pt}){4-5}
%     \cmidrule(l{6pt}r{6pt}){6-7}
%     \cmidrule(l{6pt}r{6pt}){8-9}
%  Method & P-jaw & Suction  & P-jaw & Suction  & P-jaw & Suction  & P-jaw & Suction \\
%     \midrule
%     % pi0                                      &  0.05& --     &  --    &  -- &  --    &  --   \\
%     Pi0.5                                        &  -- & --     &  --    &  -- &   &    &       &       \\
%     Pi0.5-FT                                        &  -- & --     &  --    &  -- &   &    &       &       \\
%     % pi0-GVLA                                      &  -- & --     &  --    &  -- &  --    &  --   \\
%     \rowcolor{gray!15}
%     GVLA (Ours)                                &    &      &      &    &  &  &   &     \\
%     \bottomrule
%   \end{tabular}
%   }
%   \label{tab:real-world}
%   \vspace{-0.5em}
% \end{table}

\begin{wrapfigure}{r}{0.39\textwidth}
\vspace{-5ex}
    \centering
    \includegraphics[width=0.39\textwidth]{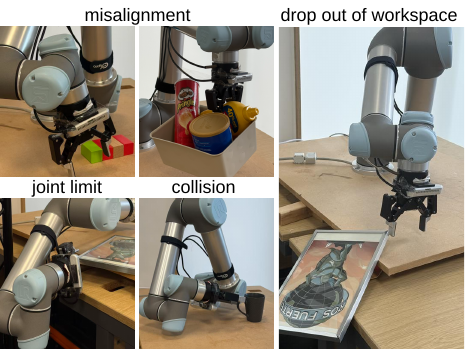}
    \vspace{-4ex}
    \caption{Failure cases}
    \label{fig:your_label}
    \vspace{-6ex}
\end{wrapfigure}

% \begin{wrapfigure}[t]
%     % \centering
%     % \includegraphics[width=0.5\linewidth]{figures/failure_cases.pdf}
%     \includegraphics[width=0.4\linewidth]{example-image}
%     \vspace{-2ex}
%     \caption{failure cases - TBD. \textbf{WONT HAVE ENOUGH SPACE FOR THIS. NEED TO COMBINE WITH OTHER FIGURE, OR MAKE IT SMALLER. PUT MORE RESULTS TO THE SUP. MATERIAL} }
%     \vspace{-4ex}
%     \label{real-workd}    
% \end{wrapfigure}

% \noindent\textbf{Experimental Setup} 

\noindent\textbf{Failure Cases.} Our experiment also reveals several failure cases that are worth noticing. Fig.~\ref{fig:your_label} shows some of these failure cases. We observe that the failure cases are mostly because of:
(\textit{i}) Intra-type misalignment: When adapting to unseen grippers, the model selects a type-consistent strategy but fails to precisely align the end-effector, exposing the lack of fine-grained geometric detail for contact planning.
% (2) Cross-domain failure: Under visually distinct domains without finetuning, gripper conditioning weakens, revealing residual entanglement between morphology and scene representations.
(\textit{ii}) Physical gripper limitation: On some tasks, although the model generates feasible trajectories, the grippers are not able to grasp the object due to the physical constraints of the gripper.
(\textit{iii}) Kinematic failure: In complex grasping tasks, the robot occasionally enters kinematically infeasible configurations without recovery, exposing a disconnect between high-level action prediction and low-level kinematic feasibility.

% 1. with unseen gripper, the robot can choose the same stratigies as same type but failed most on can't align accurately especially when the grasping need prepush, that is from in lack of fine grined robot geomerty representation, for further better generalization intra same gripper type we need to explore better way inject geometry info/ or better represent across same gripper type.
% 2. we fail to genralize the gripper into new domain, which indicate the lack of disentanglement between gripper-specific and visual representations.
% 3. in some unseen task gripper will still use topdown grasping, which indicate we need better link between object feature and  gripper embodiment.
% 4. when operating the complex task, the robot failed get stuck when close to singularity position and doesn't come back. to improve the grasping skill with vla we need better control protection the recover.  

% Performance degrades on fine-grained tasks requiring high accuracy. We observe failures in (i) thin object manipulation, where grasp contact points lack precision, and (ii) trajectory planning that occasionally drives the manipulator into kinematic singularities, causing catastrophic failures. These issues suggest current action tokenization granularity and the absence of explicit kinematic constraints limit deployment in high-precision scenarios.
% \subsection{Generalization and Adaptation}
\vspace{-3ex}
\section{Discussion}
\vspace{-3ex}
\textbf{Limitations.}
While our work shows encouraging results, it faces certain limitations. First, the simulation environment of our MiGA dataset is based on NVIDIA Isaac Lab, which has certain limitations in modeling accurate gripper motion in complex cases (e.g., soft or multi-DoF grippers). Second, in our GVLA, the gripper information is currently encoded to guide the strategy-level action differences, without explicit kinematic or contact modeling, which limits precise intra-type adaptation. Incorporating richer geometric representations could enable instance-aware contact planning. Furthermore, although our dual MoA encourages gripper-specific specialization, gripper and visual representations remain partially entangled, causing the policy to over-rely on image information under domain shift. Strengthening morphology-aware reasoning and vision disentanglement are therefore critical for robust cross-domain transfer.
%Third, the relationship between object feature and embodiment constraints is implicit, leading fallback to generic strategies on unseen tasks; tighter perception–action coupling would improve embodiment-aware generalization. 

\noindent\textbf{Conclusion.} To address the limitation of gripper invariance in existing VLAs and the gripper-dependent divergence of grasping strategies across different gripper types, we made two key contributions. First, we introduce MiGA, a multi-gripper dataset collected with five gripper types and 103,000 demonstrations from simulation and real-world robots. MiGA is designed to reveal distinct action strategies across grippers. Second, we propose GVLA, a gripper-aware VLA framework that uses our new multi-gripper tokenizer and a dual MoA design to condition the policy with action and gripper routing information. Intensive experimental results show that GVLA can effectively learn gripper-discriminative representations, outperforming recent methods in complex grasping tasks and enabling stronger few-shot adaptation to new objects, tasks, or grippers. %Our code and dataset will be released to support future research.

% This work also highlights two directions for further development. (1)Zero-shot transfer to out-of-distribution grippers on unseen platforms remains limited, as visual features can dominate morphology signals; Future work should pursue stronger disentanglement between visual and gripper-specific representations for more explicitly reasoning based on gripper morphology rather than scene appearance. (2) 
% Beyond strategy-level divergence, consistent instance-level performance requires finer modeling of gripper geometry (e.g., size, contact surface, kinematics) to enable instance-aware contact for more accurate manipulation.

%Our code and dataset will be released for future research. 

%We hope both our dataset and method can foster future research in language-driven grasp detection.

% \section*{Acknowledgements}
% Please insert your acknowledgments here.

% ---- Bibliography ----
%
% BibTeX users should specify bibliography style 'splncs04'.
% References will then be sorted and formatted in the correct style.
%
\bibliographystyle{splncs04}
\bibliography{main}

% \newpage
% \input{supplementary}
\end{document}